\documentclass[letterpaper, 10pt, conference]{ieeeconf}      %
\IEEEoverridecommandlockouts                              %
\usepackage{fontspec}
\usepackage{unicode-math}
\defaultfontfeatures{Ligatures=TeX}
\usepackage{graphics} %
\usepackage[capitalise]{cleveref}
\usepackage{tikz}
\usepackage{csquotes}
\usepackage[inkscapelatex=false]{svg}
\usepackage{tikz-3dplot}
\usepackage{pgfplots}
\usepackage{pgfplotstable}
\usepackage{etoolbox}
\pgfplotsset{compat=1.18}
\usetikzlibrary{patterns,arrows.meta, positioning,
shapes.geometric,fit,calc,3d,matrix,backgrounds}
\newcommand{\drawcube}[3]{
  \begin{scope}[shift={(#1,#2,#3)}]
    \def\L{1}
    \draw (0,0,0) -- (\L,0,0) -- (\L,\L,0) -- (0,\L,0) -- cycle;
    \draw (0,0,0) -- (0,0,\L);
    \draw (\L,0,0) -- (\L,0,\L);
    \draw (\L,\L,0) -- (\L,\L,\L);
    \draw (0,\L,0) -- (0,\L,\L);
    \draw (0,0,\L) -- (\L,0,\L) -- (\L,\L,\L) -- (0,\L,\L) -- cycle;
  \end{scope}
}
\tikzset{
  voxelmap/.pic={
    \drawcube{0}{0}{0}
    \drawcube{1}{0}{0}
    \drawcube{0}{1}{0}
    \drawcube{0}{0}{1}
  }
}

\tikzset{
  stackedrect/.pic={
    \pgfmathsetmacro{\xsize}{3}
    \pgfmathsetmacro{\ysize}{2}
    \foreach \i in {2,1} {
      \draw[black, draw,  fill=white] (\i*0.1, -\i*0.1) rectangle
      ++(\xsize,\ysize);
    }
    \draw[black,draw,   fill=white] (0,0) rectangle ++(\xsize,\ysize);
    \node[align=center, text width=\xsize cm] at (\xsize / 2, \ysize /2) {#1};
  }
}

\tikzset{
  smallgraph/.pic={
    \pgfmathsetmacro{\xspacing}{0.5}
    \pgfmathsetmacro{\yspacing}{0.5}

    \node[circle, draw, inner sep=2pt] (a) at (0,0) {};

    \node[circle, draw, inner sep=2pt] (b) at (-\xspacing,-\yspacing) {};
    \node[circle, draw, inner sep=2pt] (c) at (\xspacing,-\yspacing) {};

    \node[circle, draw, inner sep=2pt] (d) at (-1.5*\xspacing,-2*\yspacing) {};
    \node[circle, draw, inner sep=2pt] (e) at (-0.5*\xspacing,-2*\yspacing) {};
    \node[circle, draw, inner sep=2pt] (f) at (0.5*\xspacing,-2*\yspacing) {};
    \node[circle, draw, inner sep=2pt] (g) at (1.5*\xspacing,-2*\yspacing) {};

    \draw[->] (a) -- (b);
    \draw[->] (a) -- (c);

    \draw[->] (b) -- (d);
    \draw[->] (b) -- (e);

    \draw[->] (c) -- (f);
    \draw[->] (c) -- (g);
  }
}

\tikzset{
  server/.pic={
    \path[draw=black, fill=exist] (0,0) rectangle (2,3.5);
    \foreach \y in {3.0, 2.7, 2.4} {
      \draw[black, line width=1pt] (0.3,\y) -- (1.7,\y);
    }
  }
}
\tikzset{database/.style={cylinder,aspect=0.5,draw,rotate=90,path picture={
      \draw (path picture bounding box.160) to[out=180,in=180] (path
        picture bounding
      box.20);
      \draw (path picture bounding box.200) to[out=180,in=180] (path
        picture bounding
      box.340);
}}}

\usepackage{tabularx}
\usepackage{multirow}
\usepackage{booktabs}
\usepackage[acronym]{glossaries}
\glsdisablehyper
\usepackage{flushend}
\usepackage{censor}
\usepackage[svgnames,dvipsnames,table,rgb]{xcolor}
\usepackage{mwe}
\usepackage[caption=false]{subfig}
\definecolor{cornflowerblue}{HTML}{6494EC}
\definecolor{instanceNodeColor}{HTML}{FED600}
\definecolor{metaNodeColor}{HTML}{FE391D}
\definecolor{plotblue}{RGB}{93,121,173}
\definecolor{plotorange}{RGB}{218,140,93}
\definecolor{plotgreen}{RGB}{124,180,113}
\StopCensoring
\usepackage[
  backend     = biber,
  style       = ieee,
  seconds     = true,
  natbib      = true,           %
  sorting     = none,           %
  maxbibnames = 6,           %
  bibencoding = utf8,
  maxcitenames= 1,
  mincitenames= 1
]{biblatex}
\DeclareNameAlias{author}{family-given}
\AtEveryBibitem{%
  \clearfield{keywords}%
  \clearfield{note}%
}
\AtEveryBibitem{%
  \ifboolexpr{
    test {\ifentrytype{inproceedings}}
    or
    test {\ifentrytype{proceedings}}
    or
    test {\ifentrytype{article}}
  }{%
    \clearfield{url}%
    \clearfield{urldate}%
    \clearfield{urlyear}%
    \clearfield{urlmonth}%
    \clearfield{urlday}%
  }{}%
}
\AtEveryBibitem{%
  \iffieldundef{doi}{}{%
    \clearfield{url}%
    \clearfield{urldate}%
    \clearfield{urlyear}%
    \clearfield{urlmonth}%
    \clearfield{urlday}%
    \clearfield{eprint}
    \clearfield{eprinttype}
  }%
}
\AtBeginBibliography{\footnotesize}
\title{\LARGE \bf
  From Pixels to Concepts: Growing Rich 3D Semantic Scene Graph
  Forests utilizing Foundation Models
}
\author{
  \xblackout{David Oberacker}$^{1}$ \and
  \xblackout{Meike Deitersen}$^{1}$ \and
  \xblackout{Niklas Spielbauer}$^{1}$ \and
  \xblackout{Tristan Schnell}$^{1}$ \and
  \xblackout{Georg Heppner}$^{1}$ \and
  \censor{Arne Roennau}$^{1,2}$
  \thanks{
    $^{1}$\xblackout{
      David Oberacker, Meike Deitersen, Niklas Spielbauer, Tristan
      Schnell, Georg Heppner, and Arne Roennau are with the FZI
      Research Center for Information Technology, 76131 Karlsruhe, Germany.
      \texttt{oberacker@fzi.de}
    }
  }
  \thanks{
    $^{2}$\xblackout{
      Arne Roennau is with the Machine Intelligence and Robotics Lab
      (MaiRo), Karlsruhe Institute for Technology (KIT), Karlsruhe, Germany.
    }
  }%
  \thanks{
    \xblackout{
      The research leading to these results has received funding in the ROBDEKON II project under the grant agreement No. 13N16540 by the German Federal Ministry of Education and Research (BMBF) and the AIDeCo project, which is supported by donations from the Intel Corporation. 
    }
  }
}

\newcommand\copyrighttext{%
	\footnotesize \textcopyright \the\year{} IEEE. Personal use of this material is permitted.
	Permission from IEEE must be obtained for all other uses, in any current or future
	media, including reprinting/republishing this material for advertising or promotional
	purposes, creating new collective works, for resale or redistribution to servers or
	lists, or reuse of any copyrighted component of this work in other works.}
\newcommand\copyrightnotice{%
	\begin{tikzpicture}[remember picture,overlay]
		\node[anchor=south,yshift=10pt] at (current page.south) {\fbox{\parbox{\dimexpr\textwidth-\fboxsep-\fboxrule\relax}{\copyrighttext}}};
	\end{tikzpicture}%
}

\begin{document}
\newacronym{hssg}{3D-HSSG}{3D Hierarchical Semantic Scene Graph}
\newacronym{ssg}{3D-SSG}{3D Semantic Scene Graph}

\newacronym{vlm}{VLM}{Vision Language Model}

\newacronym{llm}{LLM}{Large Language Model}
\newacronym{slam}{SLAM}{Simultaneous Localization and Mapping}
\newacronym{lidar}{LiDAR}{Light Detection and Ranging}
\newacronym{nn}{NN}{Neural Network}
\newacronym{cnn}{CNN}{Convolutional Neural Network}
\newacronym{rnn}{RNN}{Recurrent Neural Network}
\newacronym{fnn}{FNN}{Feedforward Neural Network}
\newacronym{moe}{MoE}{Mixture of Experts}
\newacronym{iou}{IoU}{Intersection over Union}
\newacronym{ros}{ROS}{Robot Operating System}
\newacronym{fzi}{FZI}{FZI Research Center for Information Technology}
\newacronym{ui}{UI}{User interface}
\newacronym{gnn}{GNN}{Graph Neural Network}
\newacronym{tp}{TP}{True Positive}
\newacronym{fp}{FP}{False Positive}
\newacronym{fn}{FN}{False Negative}
\newacronym{gpu}{GPU}{Graphics Processing Unit}
\newacronym{ros2}{ROS 2}{Robot Operating System 2}

\maketitle
\thispagestyle{empty}
\pagestyle{empty}
\begin{abstract}
  Operating in complex real-world environments requires robots to understand their surroundings on a functional semantic level.
This demands a detailed multi-layer world model capturing the complex relations of the robot's surroundings.
Hierarchical 3D scene graphs address this challenge by integrating geometric, semantic, and relational data within a unified spatial framework.
However, current 3D scene graph approaches often restrict themselves to rigid structures of predetermined relationship classes, mostly neglecting important semantic connections, like causal connections or environmental contexts.
This paper explores the potential of foundation models to build forests of 3D scene graphs with open semantic relationships to improve scene understanding and robotic task execution.
We propose a method where instance-specific concept nodes and relationships are first identified by a \acrshort{vlm} and extended upon by an \acrshort{llm}, inferring broader, more abstract concept nodes and relationships through reasoning.
These object nodes, concept nodes, and relationships are then assembled into a forest of hierarchical 3D scene graphs, enhanced with concept nodes to represent abstract concepts.
Evaluations were conducted on the uHumans2 and ScanNet indoor datasets, validating the accuracy and relevance of the generated relationships.
Downstream suitability of scene-graph forests for robotics applications is demonstrated in an open-vocabulary object-retrieval task utilizing both ScanNet data and a real-world indoor deployment using a Boston Dynamics Spot.
This paper leverages foundation models to create more expressive, semantically deep 3D hierarchical scene graphs and demonstrates their potential to advance semantic and environmental understanding in robotics.

\end{abstract}
\copyrightnotice
\begin{figure}[ht]
  \centering
  \includegraphics[width=\linewidth]{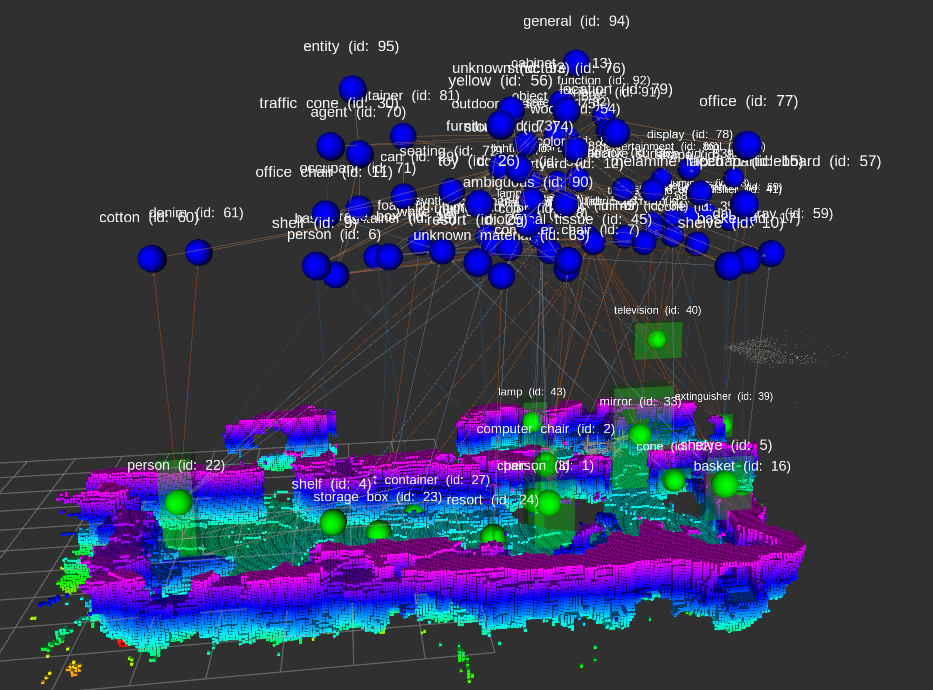}
  \caption{Generated scene graph from the real-world deployment on a BostonDynamics Spot robot together with the 3D voxel map.
    Detected objects are represented with their bounding boxes (green).
    Above the object nodes, the hierarchy of concept-nodes is shown.
  }
  \label{fig:scene_graph_visualization}
\end{figure}
\section{Introduction}
\label{sec:introduction}
Information is the foundation of any meaningful interaction or problem-solving.
We as humans continuously and subconsciously observe our environment, gathering semantic information such as objects or possible affordances.
This continuously updated model of our environment allows us to solve novel tasks faster as we are already primed with an encompassing context of our surroundings.
With mobile robots becoming more suitable for applications in human-centered environments, a similar need for a semantically deep and context-rich model of the environment arises.
Semantic 3D scene graphs have established themselves as a powerful method for modeling environments~\cite{rana_sayplan_2023,hughes_hydra_2022,maggio_clio_2024}.
While spatial representation and semantic understanding have been highly relevant topics for many years, the recent advances in embodied AI have drastically altered the requirements for semantic environment representations.
Especially the rigid, manually crafted ontologies underlying many earlier 3D scene graph perception approaches~\cite{armeni_3d_2019, hughes_hydra_2022, rosinol_kimera_2021} are unsuitable to model the complex interactions modern robots are expected to perform.
To this end, research on open-concept and open-vocabulary 3D scene graph generation has become a major focus~\cite{koch_open3dsg_2024, chen_leveraging_2023,gu_conceptgraphs_2024,chang_comprehensive_2023,chen_clip-driven_2024,chang_context-aware_2023,maggio_clio_2024}.
Approaches like ConceptGraphs~\cite{gu_conceptgraphs_2024} utilize open-vocabulary segmentation to detect objects in RGB-D images as nodes for the scene graph.
Pairwise edges between these nodes are then proposed by a~\gls{vlm} given each object's 2D pose and vision mask.
Similarly, Open3DSG~\cite{koch_open3dsg_2024} generates scene graphs directly from point cloud information utilizing a graph neural network and CLIP~\cite{radford_learning_2021}.
Whereas the aforementioned approaches focus mostly on spatial relationships,~\citeauthor{zhang_open-vocabulary_2025}~\cite{zhang_open-vocabulary_2025} aim at identifying functional relationships, namely affordances, between object pairs.
Opposed to earlier approaches, the majority of open-vocabulary systems forgo hierarchical structure in favor of denser graphs.
\par
Utilizing the strong contextual knowledge of~\glspl{llm} has proven to provide significant benefits for generating open-vocabulary 3D scene graphs~\cite{zhang_open-vocabulary_2025, wu_scenegraphfusion_2021, chen_clip-driven_2024, koch_open3dsg_2024,koch_lang3dsg_2024}.
When looking at the larger problem of generating encompassing environment models that are suitable for use in complex multi-step tasks, some open issues remain.
First, current approaches often restrict themselves to singular relationship types (affordances, spatial), while utilizing open-vocabulary node and edge labels.
A second open question is multi-level hierarchies, where more than just pairwise connections between object nodes are proposed.
Both are required to provide the robot with similar information depth as a human would acquire naturally.
\par
This work provides an approach for building a set of hierarchical 3D scene graphs in parallel, representing different relationship types and concept levels.
\begin{itemize}
  \item We introduce an approach for building an information-dense \textbf{forest of~\gls{hssg}} (see~\cref{fig:scene_graph_visualization}), where semantically disjunct \textbf{concept layers} provide semantic abstractions above a common \textbf{object node} layer, grounded in the scene.
  \item We propose a method to generate the aforesaid forests in a fully \textbf{open-vocabulary} manner by iteratively utilizing \glspl{llm} and \glspl{vlm} with strong \textbf{self-validation} to generate new nodes and edges based on perceived objects and previously generated contextual nodes in the scene graph.
  \item We provide a comprehensive evaluation of the generated relation quality, \textbf{validating the accuracy and consistency} of the relations and concept nodes produced by our approach.
  \item We demonstrate the applicability to \textbf{downstream robotics tasks} for \textbf{forests of~\gls{hssg}} by realizing an open-vocabulary object retrieval task in a real-world indoor deployment.
\end{itemize}
\par
First,~\cref{sec:related_works} highlights some related works.
In \cref{sec:method}, we propose the scene graph forest generation method, with the implementation in~\cref{sec:implementation}.
\Cref{sec:evaluation} analyzes the quality of generated relations and the downstream applicability of our approach.
\Cref{sec:discussion} discusses the results, outlining the advantages and limitations of our approach.
Lastly, in~\cref{sec:conclusion}, we conclude and give an outlook of future improvements.

\section{Related Works}
\label{sec:related_works}
\subsection{Closed-Vocabulary 3D Scene Graphs}
Initially proposed by~\citeauthor{armeni_3d_2019}~\cite{armeni_3d_2019} hierarchical 3D scene graphs have been well established to represent a robot's environment.
Approaches like KIMERA by~\citeauthor{rosinol_kimera_2021}~\cite{rosinol_kimera_2021} have shown that 3D scene graphs are suitable for SLAM applications, including the representation of dynamic agents, like humans or other robots.

Both approaches employ a fixed ontology, based on objects (fixed object classes), places, rooms, and buildings, for defining the layers of the scene graph.
Similarly, graph-neural-network-based approaches~\cite{wu_scenegraphfusion_2021, wald_learning_2020}, which use trained neural networks to predict 3D scene-graphs from input data, utilize static lexical onologies to organize nodes and edges in the graph.
\subsection{Open-Vocabulary 3D Scene Graphs}
To address the problem of fixed ontological models when creating scene representations, recent works have shifted towards open-vocabulary approaches for labeling nodes and edges~\cite{koch_open3dsg_2024,maggio_clio_2024,werby_hierarchical_2024,gu_conceptgraphs_2024,devarakonda_orionnav_2024}.
\par
ConceptGraphs, introduced by~\citeauthor{gu_conceptgraphs_2024}~\cite{gu_conceptgraphs_2024}, utilize streams of RGB-D data to generate 3D scene graphs iteratively.
Observed objects are first detected in an input image using an open-vocabulary approach.
Afterward, they are associated with the object in the generated scene representation using a feature-level representation.
A~\gls{llm} then generates relationships between pairs of objects utilizing label and spatial information from the 3D map.
This differs from our approach, as they primarily focus on pairwise semantic connections between objects, whereas we do not consider object-to-object connections, only connections between objects and abstract concepts.
\par
\citeauthor{koch_open3dsg_2024}~\cite{koch_open3dsg_2024} propose a learned approach that can predict a static 3D scene graph from point cloud data with both open-vocabulary nodes and edges.
During inference, a~\gls{gnn} produces a 3D graph structure from pointcloud data, where nodes are labeled by comparing the cosine similarity between CLIP~\cite{radford_learning_2021} embeddings and the node embeddings produced by the~\gls{gnn}.
Graph Edges are in turn predicted by combining an~\gls{llm} and Qformer.
One downside of this approach is that nodes in the generated graph structure are always related to objects in the scene; thus, relationships can only be expressed between pairs of objects.
Similarly to ConceptGraphs, this differs from our approach, as they focus on pairwise object relations.
\par
CLIO~\cite{maggio_clio_2024}, proposed by~\citeauthor{maggio_clio_2024}, builds open-set 3D scene graphs on top of the places graph from Hydra~\cite{hughes_hydra_2022}.
CLIP~\cite{radford_learning_2021} and Segment Anything~\cite{kirillov_segment_2023} are used to mask and identify objects in the scene, with the weighted CLIP feature vector of all nearby objects being assigned to places in the scene graph.
Opposed to our approach, CLIO specializes in task-specific querying based on object clusters, not the generation of semantic hierarchies.
\par
HOV-SG~\cite{werby_hierarchical_2024} and OrionNav~\cite{devarakonda_orionnav_2024} both focus on creating hierarchical 3D scene graphs for navigation/task planning.
In both approaches, feature maps from an open-vocabulary semantic segmentation network are projected into 3D space, and object clusters are determined in 3D.
While the nodes and node-level connections are created in an open-vocabulary manner, the hierarchy of scene graph levels is, in contrast to our approach, fixed to a model of objects, places, rooms, and buildings.
\par
A recent approach by~\citeauthor{zhang_open-vocabulary_2025}~\cite{zhang_open-vocabulary_2025} extracts~\gls{ssg} with functional dependency edge labels between nodes.
Instead of describing spatial relationships, functional~\gls{ssg} focus on interactable objects and their causal relationships (e.g. water faucet controls the water flow to the sink).
While utilizing similar approaches to edge and node generation to our approach, relying directly on~\gls{vlm} and~\gls{llm}, they do not consider hierarchical relationships and non-object nodes.
\par
Other than the presented state-of-the-art, our approach aims to grow multiple~\gls{hssg} with open-vocabulary concepts and organize them in a structured and machine-interpretable manner.

\section{Method}
\label{sec:method}
By extending the traditional \acrfull{hssg} generation~\cite{armeni_3d_2019}, from a singular graph, to a set of multiple graphs, we aim to increase the amount of information that
can be encoded in a robot's environmental model.
We propose \textbf{a forest of \gls{hssg}} where leaf nodes encode objects grounded in the scene, and inner nodes represent open-vocabulary concepts.
Each graph in the forest represents a conceptual hierarchy, where a high-level semantic concept is the root, and each leaf represents an object, and inner nodes represent sub-concepts.
Specifically, we define the forest $\mathcal{F}$ as:
\begin{equation}
  \begin{split}
    \mathcal{F} = \{(\mathcal{O},\mathcal{C}, \mathcal{R}) | &
      \mathcal{O}=~\text{object nodes} \\
      & \mathcal{C}=~\text{concept nodes} \\
    & \mathcal{R}=~\text{relationships}\}
  \end{split}
\end{equation}
Object nodes $\mathcal{O}$ encode physical objects in the scene, and ground the resulting graph in the described environment:
\begin{equation}
  \begin{split}
    \mathcal{O} = \{(i,c,p,h,w) \mid &i=~\text{unique id},\\
      &c=~\text{class label},\\
      &p=~\text{B.Box Center 6D-Pose},\\
      &h=~\text{height},\\
    &w=~\text{width} \}
  \end{split}
\end{equation}
We utilize bounding boxes for spatial grounding in relation to an additional 3D voxel map of the environment.
This avoids information duplication between the spatial and semantic models.
$\mathcal{C}$ describes concept nodes, which describe non-embodied concepts within the scene.
\begin{equation}
  \mathcal{C} = \{(l)| l=~\text{concept label}\}
\end{equation}
Lastly, $\mathcal{R}$ describes the relationships in the graphs.
\begin{equation}
  \begin{split}
    \mathcal{R} = \{(s, e, t) | & s \in \mathcal{O} \cup \mathcal{C},
      e \in \mathcal{C},\\
    & t=~\text{relationship type}\} \\
    \text{with}~\forall t : \mathcal{R}^+ & \cap \{(n,n,t) | n \in
    \mathcal{O} \cup \mathcal{C} \} = \emptyset
  \end{split}
  \label{eq:relationships}
\end{equation}
Instead of direct relationships between object nodes, we propose an infinite hierarchy of concept nodes, where direct object-to-object relationships can be represented through transitive connections via concept-nodes, e.g., \enquote{\textit{doorknob opens door}} can be represented by \enquote{\textit{doorknob}} \enquote{\textit{provides}} \enquote{\textit{open}}, and \enquote{\textit{door}} \enquote{\textit{can be}} \enquote{\textit{open}}.
While unintuitive, it retains the hierarchical nature of the forest and keeps its graphs cycle-free.
Please note that $\mathcal{F}$ does not constitute a real forest, only its subset $\mathcal{F_C}$, describing the concept-node-only subset of the graph structure.
\begin{equation}
  \begin{split}
    \mathcal{F_C} = \{(\mathcal{C}, \mathcal{R_C}) | & \mathcal{R_C}
      \subset \mathcal{R}~\text{with} \\
      \mathcal{R_C} =& \{r = (s,e,t) | r \in \mathcal{R} \land s,e \in
    \mathcal{C}\}\}
  \end{split}
\end{equation}
This is by design, as $\mathcal{F_C}$ should describe distinct conceptual structures, whereas the object nodes should connect and ground these distinct structures in the scene.
\par
To address the inherent static nature of most current approaches towards open-vocabulary~\gls{hssg}, we aim to generate both object nodes $\mathcal{O}$ and concept nodes $\mathcal{C}$ fully autonomously using~\gls{llm} and~\gls{vlm} models.
The ability of the system to generate non-embodied nodes promotes the structured connection of concepts.
For example, newly introducing different colors as concept nodes and a higher-tier \enquote{\textit{color}} concept, a structured search over the entire graph's objects that emit this property becomes possible.
By incrementally extending the concept node trees, we are able to create a context-dense representation of semantic concepts grounded in the scenes and objects, making it suitable for many robotic downstream applications.
\par
\begin{figure*}[ht]
  \vspace{0.5cm}
  \centering
  \resizebox{\textwidth}{!}{%
    \begin{tikzpicture}[
		>={Stealth},
		node distance=1.5cm and 2cm,   %
		]

        \tikzset{
        	pipelineNodeStyle/.style={
        		shape=rectangle,          %
        		rounded corners,
        		draw=cornflowerblue,       %
        		fill=cornflowerblue!20,              %
        		line width=2pt,
        		inner sep=6pt,
        		align=center,
        		font=\large\sffamily,
                minimum width=4cm,
                minimum height=2.5cm
        	}
        }
        \tikzset{
            edgeNodeStyle/.style={
                line width=2pt,
                rounded corners
            }
        }

		\node[] (rgbd)      {
        \tikz {
            \pic[scale=1.0,font=\large\sffamily]{stackedrect={RGB-Images + Camera Poses}};
          }};

		\node[pipelineNodeStyle, right=of rgbd,dashed, pattern=north west lines, pattern color=cornflowerblue!20] (yolo)       {Open-Vocabulary \\ Object Detection\\{\small\textit{(YOLO-E)}}};
		\node[pipelineNodeStyle, below=of yolo, dashed,pattern=north west lines, pattern color=cornflowerblue!20] (vdb)        {vdb\_mapping};
        \node[right=of vdb] (voxelmap) {
            \tikz {
              \pic[scale=0.8,font=\large\sffamily] {voxelmap};
            }
          };
        \node[below=0.1cm of voxelmap,align=center,font=\large\sffamily] (voxelmapl) {3D Voxel Map};
		\node[pipelineNodeStyle, right=of yolo] (objdet)     {Long-Term\\Multi-Object Tracker};

		\node[pipelineNodeStyle, right=of objdet, fill=instanceNodeColor!20, draw=instanceNodeColor] (inst)    {Instance\\Concept Creation};
		\node[pipelineNodeStyle, right=of inst, fill=metaNodeColor!20, draw=metaNodeColor] (meta)      {Meta\\Concept Creation};
		\node[pipelineNodeStyle, right=of meta, fill=WildStrawberry!20, draw=WildStrawberry] (sg) {Scene Graph\\Generation};
        \node[right=of sg, text width = 3cm, align=center] (sgg) {
            \tikz{
                \pic[scale=1.5] {smallgraph};
            }
        };
        \node[below=0.1cm of sgg,align=center,font=\large\sffamily] (sggl) {Forest of\\Hierachical 3D Semantic\\Scene Graph};
		\node[pipelineNodeStyle, below=of inst, fill=OliveGreen!20,draw=OliveGreen] (vlm)	{Visual Language\\Model \\{\small\textit{(Qwen3-VL)}}};
		\node[pipelineNodeStyle, below=of meta, fill=OliveGreen!20,draw=OliveGreen] (llm)	{Reasoning\\Large Language\\Model\\{\small\textit{(Qwen3-Coder)}}};
        \begin{scope}[on background layer]
			\node[draw, dashed, line width=2pt, inner sep=15pt,rounded corners, fit=(yolo) (vdb) (voxelmap) (voxelmapl) (objdet),label={[label distance=1cm]above:{\sffamily\large\textbf{Object Node Extraction}}} ] (obj_node_ext) {};

            \node[draw,dashed, line width=2pt, inner sep=15pt,rounded corners, fit=(inst) (meta),label={[label distance=1cm]above:{\sffamily\large\textbf{Concept Node and Relationship Extraction}}} ] (concept_node_ext) {};

            \node[draw,dashed, line width=2pt, inner sep=15pt,rounded corners, fit=(sg)(sgg) (sggl),label={[label distance=1cm]above:{\sffamily\large\textbf{Scene Graph Assembply}}} ] (sg_assembly) {};

            \node[draw,dotted, line width=2pt, inner sep=15pt,rounded corners, fit=(llm)(vlm),label=below:{\sffamily\large\textbf{Compute Server}} ] (polyserve) {};

		\end{scope}

		\draw[->,edgeNodeStyle] (rgbd.east)   -- (yolo.west);
		\draw[->,edgeNodeStyle] (yolo.east)   -- (objdet.west);
		\draw[->,edgeNodeStyle] (rgbd.south)    |- (vdb.west);
		\draw[->,edgeNodeStyle] (objdet.east) -- (inst.west);
		\draw[->,edgeNodeStyle,dashed] (objdet.south) |-  ($(vdb.north) + (0,1)$) -- (vdb.north);
        \draw[<->,edgeNodeStyle] (vdb.east) -- (voxelmap.west);
		\draw[->,edgeNodeStyle] (objdet.north) |- ($(meta.north)+ (0,1)$) -- (meta.north);
		\draw[->,edgeNodeStyle] (inst.east)   -- (meta.west);
		\draw[<->,edgeNodeStyle] (inst.south) -- (vlm.north);
		\draw[<->,edgeNodeStyle] (meta.south) -- (llm.north);
		\draw[->,edgeNodeStyle] (meta.east)   -- (sg.west);
        \draw[->,edgeNodeStyle] (sg.east) -- (sgg.west);

	\end{tikzpicture}
  }
  \caption{
    High-level overview of our approach for generating a
    \textbf{forest of~\gls{hssg}} from RGB-D data and camera poses
    using a \textcolor{OliveGreen}{\gls{vlm}} and
    \textcolor{OliveGreen}{\gls{llm}}.
    The approach consists of four main stages:
    \textcolor{cornflowerblue}{(1) Semantic Object Perception};
    \textcolor{instanceNodeColor}{(2) Instance Specific Concept-Node
    Generation}; \textcolor{metaNodeColor}{(3) Meta-Level
    Concept-Node Generation}; and \textcolor{WildStrawberry}{(4)
    Hierarchical 3D Scene Graph Construction}.
    The dashed line shows the usage of the component's services.
  }
  \label{fig:pipeline_overview}
\end{figure*}
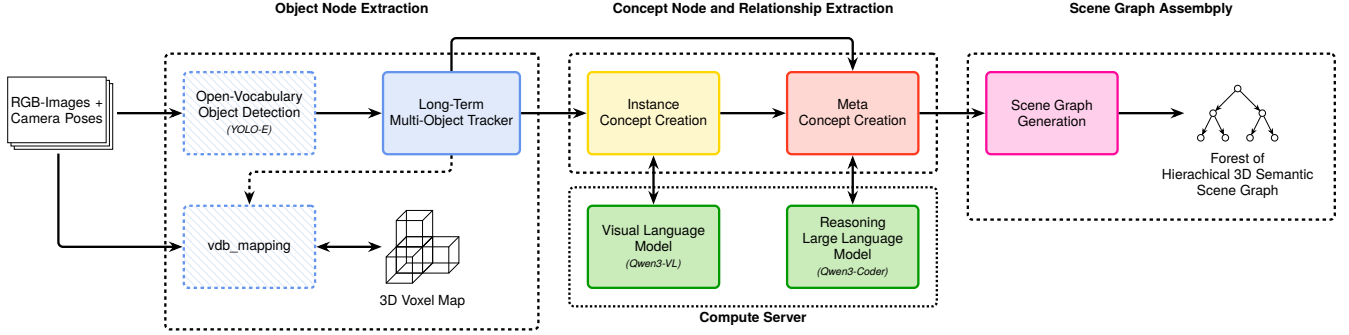
We propose a multi-step approach to incrementally create the proposed \textbf{forest of~\gls{hssg}} from a sequence of RGB-D images and 6D-poses, shown in \cref{fig:pipeline_overview}.
The framework can be separated into four distinct parts: the object node extraction, the instance concept-node generation, the meta-level concept-node generation and relationship extraction, and the scene graph assembly.
\subsection{Object Node Extraction}
First, we detect objects in the current RGB-D image utilizing existing open-vocabulary object detection methods.
At this stage, multiple detection methods are possible, given that they output a 6D pose estimate, a class estimate, and can separate instances.
For this work, we utilized YOLO-E~\cite{wang2025yoloerealtimeseeing} for this step.
Simultaneously, using \textit{vdb-mapping} a 3D-voxel map~\cite{besselmann_vdb-mapping_2021} is constructed to allow for object tracking in 3D.
6D object poses are estimated by merging overlapping detections based on their bounding box position in the 3D voxel map and the detected class label.
Each unique detection class is confirmed by the~\gls{vlm} given its proposed class and detection masks.
\subsection{Concept Node and Relationship Extraction}
After object detection, we iteratively extract concept nodes and relationships from the uniformly subsampled image stream.
As seen in~\cref{fig:pipeline_overview}, this happens in two steps: first, concept nodes, directly related to detected object nodes, are constructed to ground the context of the~\gls{vlm}/\gls{llm} to the scene in further processing steps.
Secondly, higher-level meta concept-nodes are generated based on previously generated concept-nodes.
\par
Object-specific attributes are often directly tied to the visual context.
Thus,~\glspl{vlm} visual understanding ability makes them well suited for proposing such object-specific attributes, with the main downside being that~\glspl{vlm} are prone to hallucinations.
For each frame, the input image, all object detections from the previous step, and the detection masks of objects present in the given input image are provided as inputs to a~\gls{vlm}.
It does a zero-shot prediction of attributes it can infer for detections from the image.
While we suggest the model only infer information about detections that are actually present in the image, we explicitly permit attributes for all detections to be generated, to allow for contextual interpretation, e.g., from the background.
Previously generated attributes are also provided in subsequent frames.
To guide the generation process towards common attributes, we evaluated the approach with the following relationship types in the prompt: color, material, manipulability, and semantic classes.
Others can easily be added at any time through our prompting system.
Spatial relations are explicitly ignored as a full 3D voxel map of the environment is created in parallel, which encodes these relationships with lower error thresholds than~\gls{vlm} generated information.
\par
In contrast to object-specific attributes, conceptual relationships are more dependent on advanced reasoning abilities and extensive world knowledge.
While~\glspl{vlm} provide both, recent reasoning-specialized thinking~\glspl{llm} have expressed significantly stronger performance in both fields~\cite{qwen3technicalreport}.
Therefore, we utilize a~\gls{llm} for the creation of higher-level context nodes and forgo visual context in this phase (see~\cref{fig:pipeline_overview}).
For this phase, we also provide some example relationships to the~\gls{llm}: affordances and categorical relations.
\subsection{Scene Graph Assembly}
As a last step, the generated relationship triplets~(see~\cref{eq:relationships}) are organized in a set of hierarchical scene graphs.
First, all unique nodes are generated from the start and endpoints of the relations (starting with object nodes).
Secondly, relations are added sequentially, only if they would not induce a cycle in the graphs; otherwise, they are discarded.
Together, the object nodes and concept nodes form a hierarchical structure within the 3D scene graph.
No additional rigid structure, such as an ontology, is enforced within the meta-concept layer, allowing for flexibility and ease of maintenance.
This approach ensures that the scene graphs can be easily extended with new semantic relationships.

\section{Implementation}
\label{sec:implementation}
The approach was implemented as part of the~\gls{ros2} ecosystem.
We utilized the Ultralytics YOLO-E~\cite{jocher_ultralytics_2023}~\gls{ros2} package for object detection and bounding box estimation, due to its high accuracy, open-vocabulary detections and high detection frequency.
\par
As a~\gls{vlm} we utilized Qwen3-VL~\cite{qwen3technicalreport} (8B parameters, Instruct) and as a~\gls{llm} Qwen3~\cite{qwen3technicalreport} (32B parameters, Thinking) as a thinking model.
The output of all models was post-processed and parsed into relationship triples (see~\cref{eq:relationships}) utilizing fixed parsing rules.
These rules included the removal of \texttt{<think>} tokens, or repetitions of the task before answering.
\par
The 3D-Voxel map was created using the \texttt{vdb\_mapping} framework~\cite{besselmann_vdb-mapping_2021}.

\section{Evaluation}
\label{sec:evaluation}
To validate the ability of the approach to generate novel relationships and map out complex connections between grounded object nodes and non-embodied concept nodes, we evaluated the system on the uHumans2~\cite{rosinol_kimera_2021} and ScanNet~\cite{dai_scannet_2017} datasets.
The downstream task suitability of the approach is validated through an open-vocabulary object retrieval task, evaluated on the ScanNet~\cite{dai_scannet_2017} dataset and a real world scenario with a Boston Dynamics Spot robot.
We aim to address the following research questions:
\begin{enumerate}
  \item \label{rq1} How accurately does the approach extract and generate semantic relationships?
  \item \label{rq2} How suitable are the forest of~\gls{hssg} for usage in downstream robotic tasks?
\end{enumerate}
\subsection{Generation of 3D-HSSG forests}
\begin{figure}[t]
  \centering
  \vspace*{3mm}
  \includegraphics[width=\linewidth]{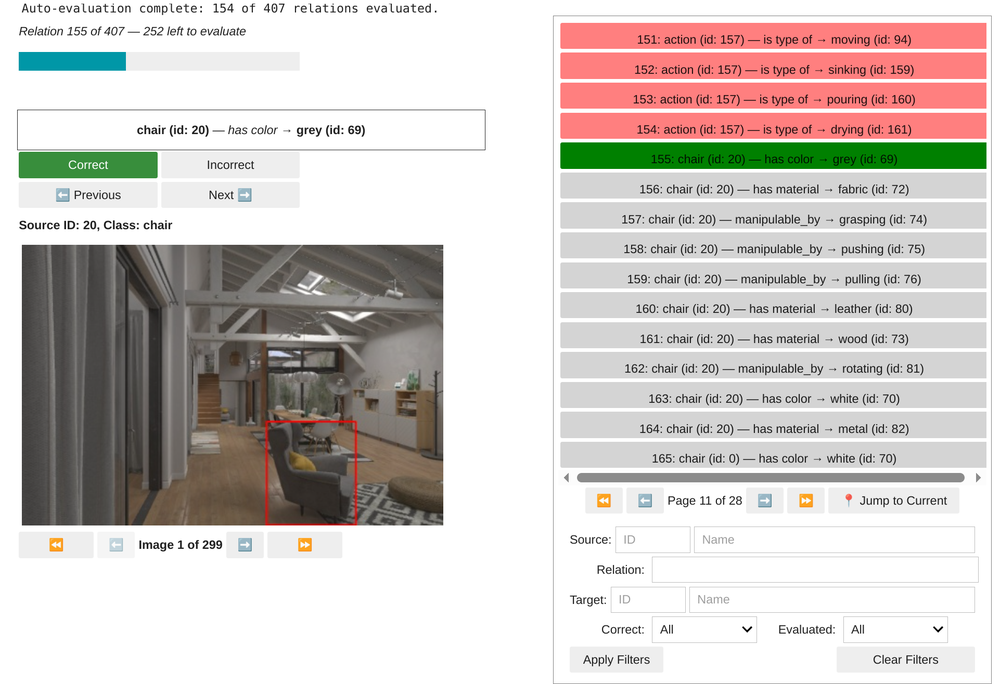}
  \caption{
    For evaluation, a Jupyter notebook-based user interface is used
    to manually evaluate generated concept-nodes and relationships.
    On the left, the image frame and generated relationship are
    displayed; the operator classifies them as correct or incorrect.
  }
  \label{fig:ui_screenshot}
\end{figure}
First, we investigated the approach's ability to generate open-vocabulary relationships.
We selected the apartment environment from the uHumans2~\cite{rosinol_kimera_2021} and a small one-bedroom apartment from the ScanNet~\cite{dai_scannet_2017} dataset as they present cluttered, human-centered environments with a significant number of possible semantically-relevant relationships.
Each dataset was processed in three runs to ensure repeatability.
Given the open-vocabulary nature of the generated relationships and concept nodes, a human-in-the-loop validation was utilized to categorize generated information as correct or incorrect, given the context of the scene.
\cref{fig:ui_screenshot} shows the custom-developed user interface for this purpose.
During the grading, we assumed the relationship type~\textit{\enquote{has class}} was always labeled as correct.
This was to account for the behavior where the corresponding object in the image, for example, a detected \textit{\enquote{toilet}}, is visually incorrect or misclassified, but correct relationships involving this object (e.g., \textit{\enquote{toilet}, \enquote{has material}, \enquote{ceramics}}) were proposed by the system.
Thus, while the basic detections from YOLO-E might be wrong, the concept node and relation in the forest were still considered correct, as the approach does not assume failures in previous processing steps.
Based on the correct and incorrect relationships, we calculated the accuracy of the models.
\begin{figure*}[t]
  \vspace*{3mm}
  \centering
  \subfloat[
    Hierarchical levels are displayed on a concentric ring showcasing the various edge types between the nodes.
    \label{fig:example_subg}
  ]{
    \includegraphics[width=0.33\textwidth]{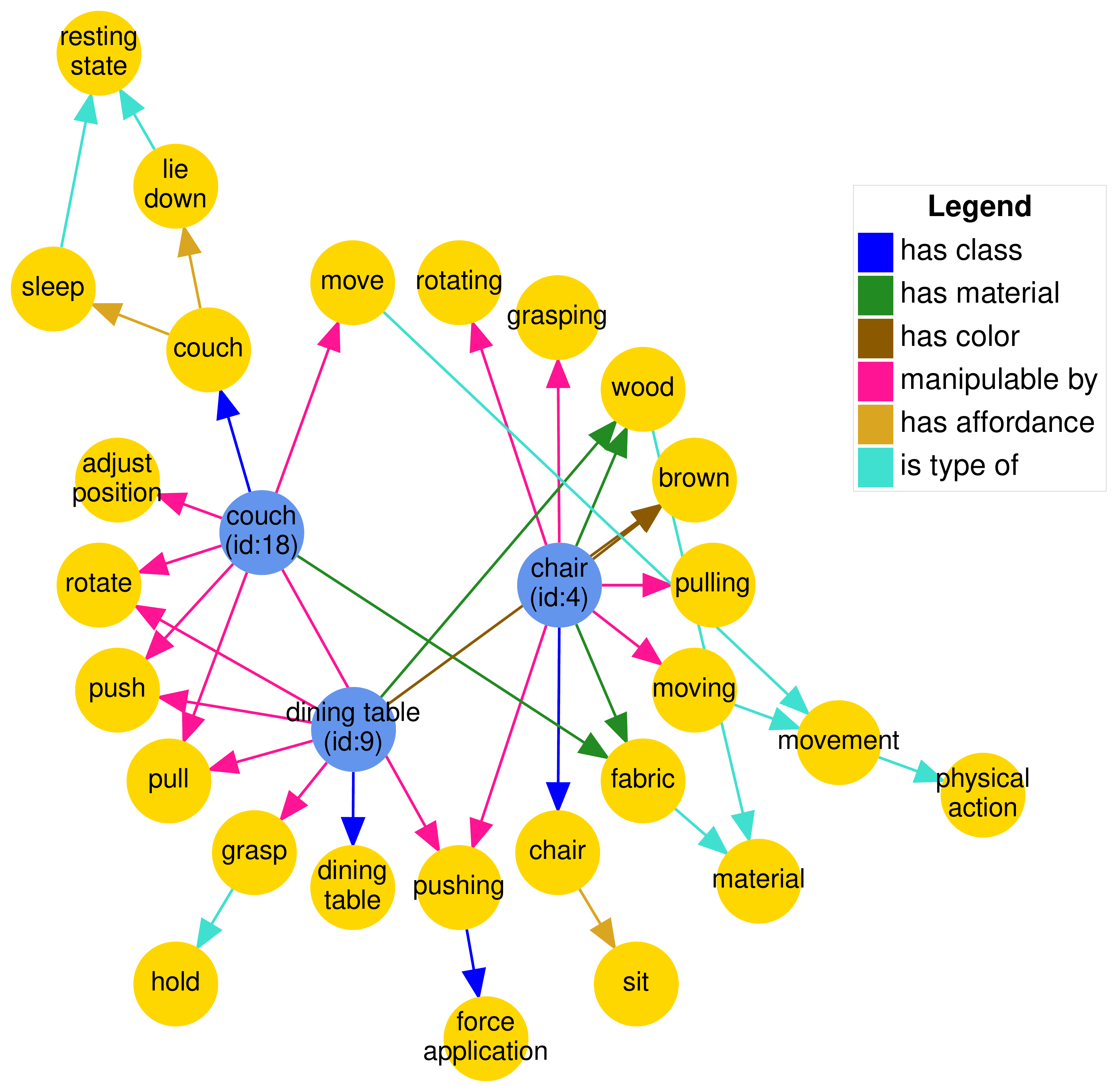}
  }
  \hspace*{2mm}
  \subfloat[
    Subset of the Semantic subgraphs depicted as distinct trees.\\
    The object-layer nodes (\textcolor{cornflowerblue}{blue}) are grounding each subtree, but are not considered connections between the different trees.
  \label{fig:example_subg_hierachy}]{
    \includegraphics[width=0.62\textwidth]{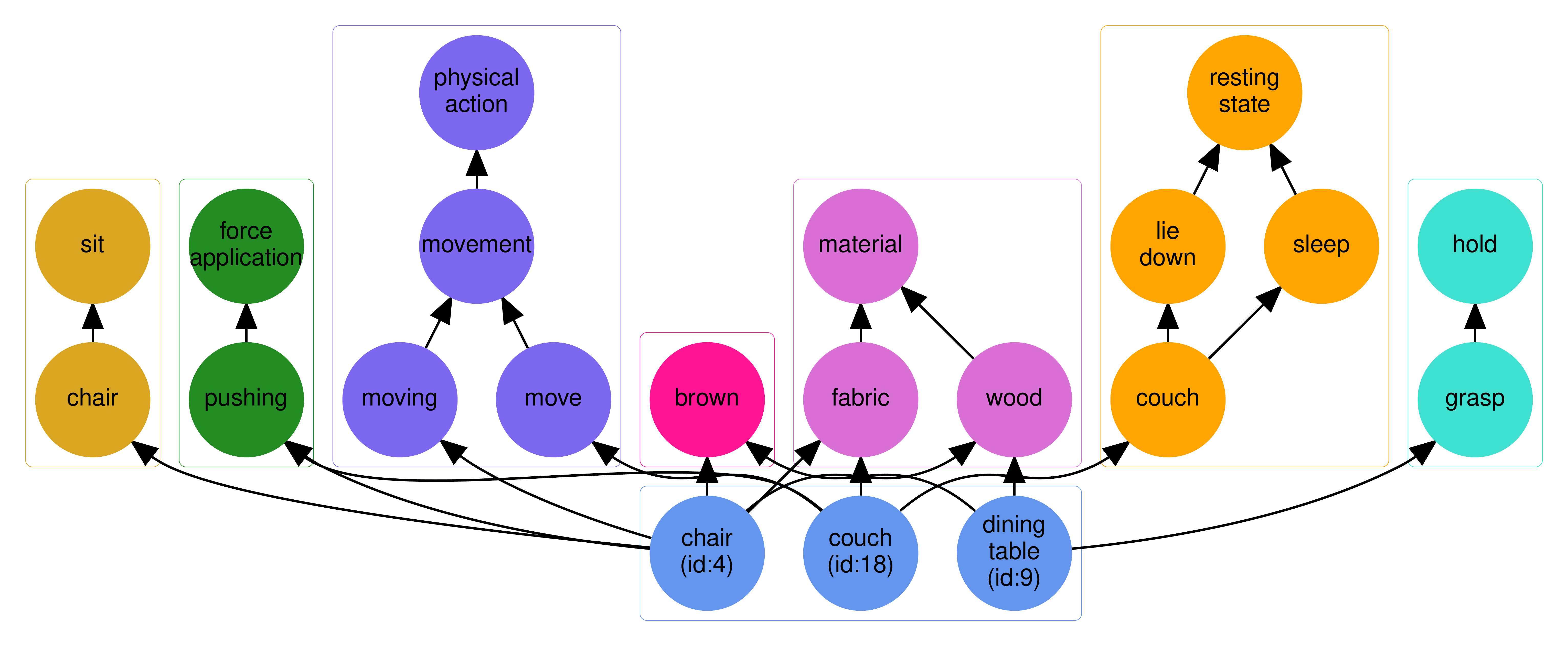}
  }
  \caption{
    Excerpt from the forest of~\acrlong{hssg} created from the uHumans2 dataset showing the hierarchies related to three objects (chair, couch, and dining table).
    For visual clarity, only one level of the subtree was included.
  }
  \label{fig:example_subg_comp}
\end{figure*}
\begin{table}[t]
  \vspace{0.5cm}
  \centering
  \caption{
    Mean and standard deviation of relationships generated, correct relationships, and accuracy (\%) per configuration and  dataset across individual runs. "Overall" rows represent aggregated results across both datasets, with statistics computed over all runs.
  }
  \label{tab:relationship_accuracy}
  \begin{tabularx}{\linewidth}{l >{\centering\arraybackslash}X >{\centering\arraybackslash}X >{\centering\arraybackslash}X}
    \toprule
    \textbf{Dataset} & \textbf{\# Relationships Generated} & \textbf{\# Correct Relationships} & \textbf{Accuracy (\%)} \\
    \midrule
    uHumans2         & $396.3 \pm 9.2$     & $261.7 \pm 6.4$    & $66.0 \pm 0.1$   \\
    ScanNet          & $529.3 \pm 168.3$   & $290.7 \pm 47.4$   & $56.8 \pm 10.3$  \\
    \textbf{Overall} & $462.8 \pm 129.1$   & $276.2 \pm 34.2$   & $\mathbf{61.4 \pm 8.2}$  \\
    \midrule
    \textbf{Relation} &  &  &  \\
    \midrule
    has class       & $228 \pm 0.0$ & $228 \pm 0.0$ & $100.00 \pm 0.0$ \\
    manipulable by  & $1002 \pm 5.2$ & $732\pm 1.7$ & $\underline{73.1 \pm 1.5}$ \\
    has material    & $462 \pm 25.9$ & $275 \pm 7.5$ & $60.2 \pm 5.7$  \\
    has color       & $419\pm21.6$ & $234\pm2.7$ & $57.0\pm11.6$  \\
    is type of      & $329\pm140.7$ & $141\pm26.2$ & $56.9\pm32.7$  \\
    has affordance  & $337 \pm 93.0$ & $47\pm13.3$  & $16.3\pm8.0$  \\
    \bottomrule
  \end{tabularx}
\end{table}
\cref{fig:example_subg} shows a subgroup of the forest that was generated from the uHumans2 apartment dataset.
The same subgroup of the forest structure, with the individual SSGs marked, is shown in~\cref{fig:example_subg_hierachy}.
The quantitative results can be seen in~\cref{tab:relationship_accuracy}.
\par
Overall, to answer research question~\ref{rq1}, the pipeline achieved a combined accuracy of \textasciitilde61.4 \% with an average of 276.2 correct relationships generated from around 50 object nodes, showing the system's ability to generate a sizable number of valid semantic relationships.
While higher accuracies (around 60-70 \%) were shown for visual contextualized relationships (is type of, has color, or has material), knowledge-dependent relationships (has affordance) had low accuracy.
The node accuracy values are comparable to similar approaches like ConceptGraphs~\cite{gu_conceptgraphs_2024}, which report a node accuracy of 61-71\%.
\par
When inspecting generated relations (see~\cref{fig:example_subg}), it can be observed that the approach struggles with lexical variations in generated concepts (e.g., grasp vs grasping, push vs. pushing, rotate vs. rotating), leading to node duplications.
On a semantic level without penalizing duplicaitons, the accuracy was therefore higher.
Another common problem was self-referential relationships, especially for the instance-level relations.
A sizable number of manipulation relations were generated; however, abstraction to meta-level relationships, like affordances, proved difficult.
The approach has shown the ability to generate various semantically sound relationships and concept-nodes (see~\cref{fig:example_subg_hierachy}).

\subsection{Downstream Task Suitabiltity}
To validate the suitability and advantages of a forest of~\gls{hssg} in downstream task applications we utilized our approach to implement an open-vocabulary object retrieval system.
We presented the generated scene graph to a Qwen3-Coder~\cite{qwen3technicalreport} (30B parameters, Mixture-of-Experts, Instruct)~\gls{llm} model with three methods — \textit{flat no attributes, flat with attributes} and \textit{graph-based} — and tasked it with identifying the objects requested by the user.
The \textit{flat no attributes} method presented only the objects in the object layer (the innermost ring shown in~\Cref{fig:example_subg_hierachy}) as a flat list to the agent.
This represents traditional scene graph methods where only object classes are recorded.
The \textit{flat with attributes} method additionally provides the related instance concept nodes for each object node.
This is what comparable semantic scene graph methods utilize, with our extension of the instance concept nodes and edges being completely autonomously defined and extracted by the~\gls{vlm} and~\gls{llm} pipeline.
Lastly, the \textit{graph-based} method provides the agent with tools, implemented in the~\texttt{smolagents}~\cite{smolagents} framework as a ReAct~\cite{yao2023reactsynergizingreasoningacting}-like reasoning system, to traverse the graph either bottom-up (from objects to concepts) or top-down (from concepts to objects).
This method utilizes the full potential of the forests of scene graphs at an increased complexity for the agent operating on the data.
\par
First, we compared the performance of these three methods for object retrieval on offline datasets.
This was done on the scene graph generated from the ScanNet~\cite{dai_scannet_2017} dataset utilized before.
The comparison aims to showcase that our fully open-vocabulary scene-graph structure is suitable for downstream tasks, with various access methods, including utilizing the full graph structure, with significant drops in task performance.
\par
Secondly, we evaluated the system in a real indoor environment with a Boston Dynamics Spot.
For this use case, unstructured natural language tasks were given to the robot, and the open-vocabulary scene graph forests were used to identify objects mentioned in the task query.
These identified objects were approximately localized using the bounding box data in the graph and then masked and grasped using previously existing components.
A key aspect of this evaluation was the disambiguation of similar objects and the utilization of conceptual information to identify the correct object candidates.
\begin{figure}
  \centering
  \subfloat[Indoor Environment for real-world evaluation]{
    \includegraphics[width=0.9\linewidth]{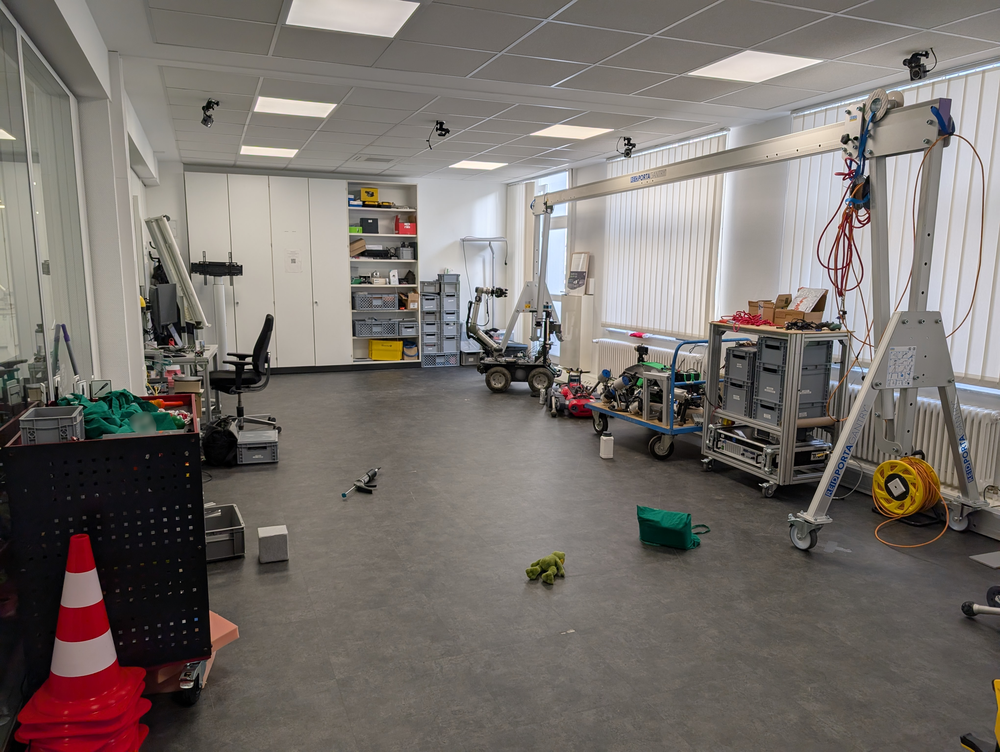}
  }
  \hfill
  \subfloat[Boston Dynamics Spot grasping an identified object]{
    \includegraphics[width=0.9\linewidth]{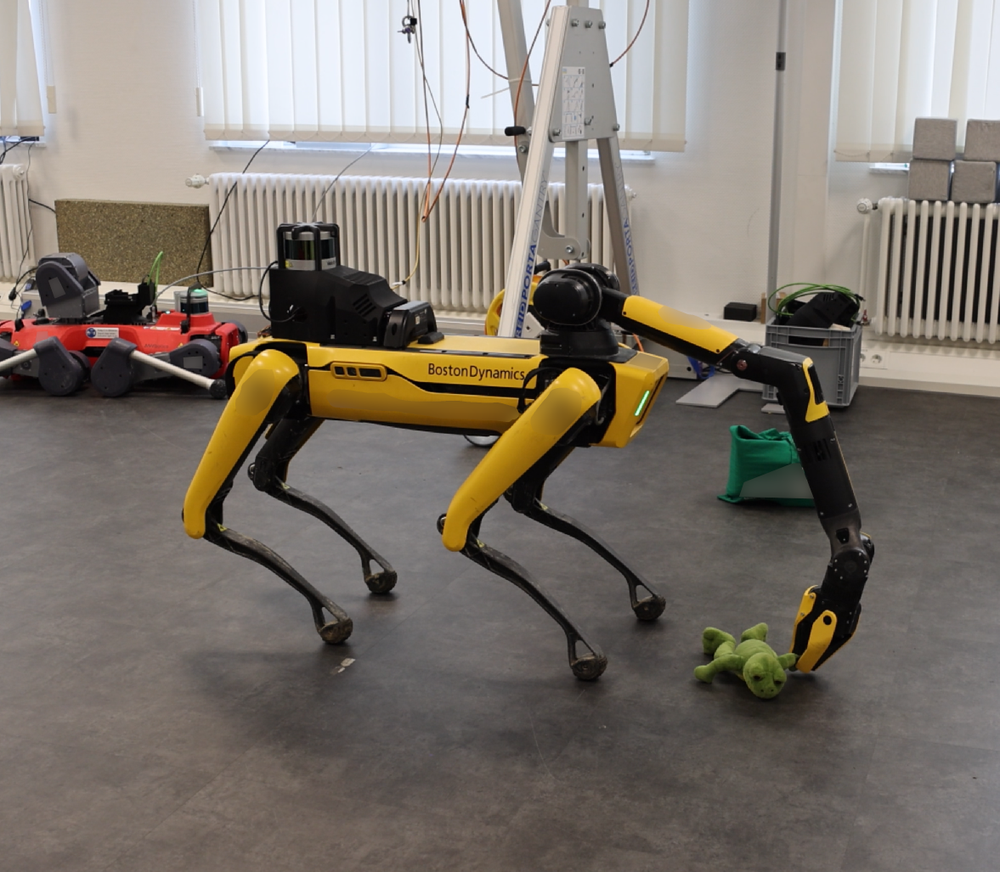}
  }
  \caption{
    Visualization of the indoor environment and BostonDynamics Spot utilized for object retrieval.
    \Cref{fig:chain_of_thought} showcases an example query to the robot from this scenario.
  }
  \label{fig:spot_openhouse}
\end{figure}
\begin{figure}
  \centering
  \vspace*{3mm}
  \includegraphics[width=0.88\linewidth]{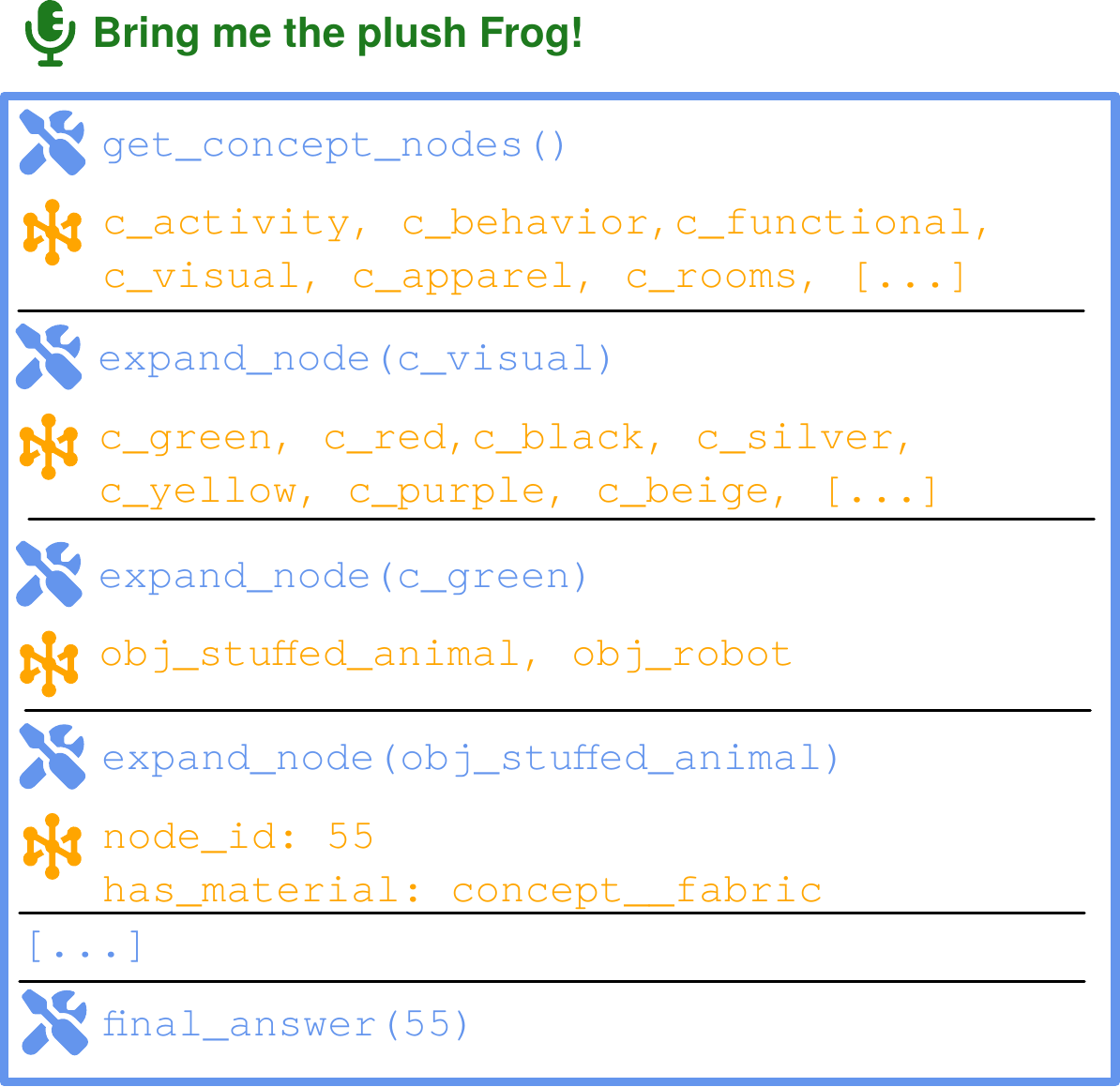}
  \caption{
    Visualization of the task query given to the robot, the object description searched for in the scene graph, and the chain-of-thought of the agent.
  }
  \label{fig:chain_of_thought}
\end{figure}
\begin{table}[t]
  \centering
  \caption{
    Object retrieval task accuracy for the three different access methods of the~\gls{hssg} forest data.
    Accuracy reports the number of times the system identified the correct object in the scene based on the query.
    Overall, three runs with 20 objects each were queried per scene.
    For the ScanNet scene, 80 objects were identified and tracked by the scene graph system.
  }
  \begin{tabularx}{\linewidth}{
      >{\raggedright\arraybackslash}X  %
      >{\raggedright\arraybackslash}X  %
      >{\centering\arraybackslash}p{2cm}  %
    }
    \toprule
    \textbf{Method} & \textbf{Environment} & \textbf{Avg. Accuracy (\%)} \\
    \midrule
    Flat no attr & \textit{ScanNet}         & 73.3 \\ %
    Flat with attr  (\textit{ours}) & \textit{ScanNet}         & 86.6 \\ %
    Graph-Based (\textit{ours}) & \textit{ScanNet}       & 70.0 \\ %
    \midrule
    Flat no attr & \textit{real}       & 80   \\ %
    Flat with attr (\textit{ours}) & \textit{real}            & 90  \\
    Graph-Based (\textit{ours}) & \textit{real}            & 100   \\
    \bottomrule
  \end{tabularx}

  \label{tab:downstream_summary}
\end{table}
\Cref{tab:downstream_summary} shows the accuracy and runtime of the object retrieval queries.
Overall, three runs of 20 queries were run with each method.
For the ScanNet dataset, more than $70\%$ of objects were disambiguated and identified correctly.
Given the high object density and strong similarities between objects (e.g. diaper bag vs. backpack vs. duffle bag), the positive impact of the generated instance-level concept nodes can be highlighted, as they increased the amount of correctly identified objects by $\approx13\%$.
The graph-based method does not significantly decrease performance compared to the baseline flat no attributes method, while the~\gls{llm} has to utilize the more complex datastructure.
As showcased in~\Cref{fig:chain_of_thought}, the active context management induced by a traversable graph structure can guide an agent to a structured thought process, reducing hallucinations and false positives.
The difference in performance for the graph-based method between the real world and ScanNet data can be explained by the increased density of the generated graph (\~100 nodes vs. \~400 nodes), resulting in the agent failing to identify relevant regions before reaching its maximum allowed steps.
This is an implementation specific restriction that does not limit the general applicability of the approach and will be addressed in future work.
These observations mostly transfer to the real-world deployment, with the performance difference being attributable to more noisy object detections resulting in less tracked objects, and therefore a higher identification rate.
Therefore, regarding research question~\ref{rq2} we find that the hierarchical structure and inclusion of concept nodes have a positive impact on the abilities of an~\gls{llm} agent to consume the environmental model for downstream task applications.
The grounding of the graph structures through the object-level nodes helps to keep the direct transfer of reasoning results to real-world actions.

\section{Discussion}
\label{sec:discussion}
Given the results presented in the previous section, we can observe that our approach is capable of producing a significant number of relevant concept-nodes given the provided relationship types.
While meta-level relations are still sparse, the instance-specific concepts generated were numerous, diverse in nature, and highly robotics-relevant.
\par
Many simple affordances related to object instances were generated effectively.
With each tree of concept nodes being rooted in a meta-level concept and grounded in the object nodes, this provides an intuitively traversable structure for finding objects adhering to specific concepts (as seen in~\cref{fig:example_subg_hierachy}).
The hierarchy, going from meta-concepts to instances, complements concept-driven approaches, like SayPlan~\cite{rana_sayplan_2023}, by exposing more suitable initial concepts with shorter graph expansions to actual object nodes grounded in reality.
This can also be observed in the agent traces shown in~\cref{fig:chain_of_thought}.
\par
Accommodating more perception-based applications, like SLAM, can be archived by adding grounded, instance-specific layers in the scene graph (e.g., places, rooms, etc.).
A conceptually interesting idea is to combine the real-time capable object-focused pipelines from semantic SLAM systems~\cite{hughes_hydra_2022} with the conceptual expansion of our approach.
For applications like semantic navigation, this will yield additional benefits, like enhanced contextual considerations for queries and more exact and up-to-date grounding of objects.
\par
As mentioned, the proposed approach shows clear benefits for its application in semantic perception.
It demonstrated its ability to generate novel and nested semantic relationships that are relevant for robotic tasks like object retrieval without requiring prior modeling.
The open-vocabulary nature of our system enables the free-flow generative approach to concept node creation.
\par
The utilized non-realtime processing approach allows for the usage of high-compute foundation models and complex query structures.
While this is required for the extensive utilization of the~\gls{vlm} and reasoning~\gls{llm}, future improvements will aim towards a continuous live processing approach, reducing the limitations for usage in mobile robotics applications, where bandwidth or network availabilty in general might be limited.
We especially want to shift towards a continuously updating data structure, as outlined in our motivation in the introduction.

\section{Conclusion}
\label{sec:conclusion}
In this work, we proposed an approach to generate a \textbf{Forest of~\acrlongpl{hssg}} directly from RGB-D images using a~\gls{vlm} and~\gls{llm} in an open-vocabulary-based approach.
Each disjunct semantic scene graph is grounded in the scene on the object layer and models a hierarchy of semantic concepts, with meta-level concepts as the root and more specific concept nodes as inner nodes (see~\cref{fig:example_subg_hierachy}).
The method leverages a multi-step process to first generate the object nodes, followed by the instance-specific non-embodied concept nodes, and lastly the meta-level concept nodes.
This approach has shown its ability to generate numerous novel, semantically correct, and robotics-relevant concept nodes and relationships in two realistic offline datasets and one real indoor scene during our evaluations.
We have shown the suitability of open-vocabulary forest of~\gls{hssg} for downstream robotic applications.
\par
In summary, our contributions are:
\begin{itemize}
  \item We define \textbf{forests of~\acrlongpl{hssg}} as a dynamic and spatially grounded semantic environment model.
  \item We show and evaluate the ability to generate novel and semantically deep scene graphs in an open-vocabulary manner by leveraging a~\gls{vlm} and \gls{llm} on two state-of-the-art indoor scene datasets.
  \item We validate the approach for downstream robotics applications through an open-vocabulary object retrieval application, both on indoor datasets and in a real-world indoor deployment.
\end{itemize}
Overall, we show the benefits of leveraging foundation models to generate complex and dynamic 3D-scene graphs with a focus on semantic environmental perception in mobile robotics use cases.
\par
Follow-up work will aim to integrate the generation process directly in real-time mapping frameworks, improve the system's ability to generate meta-level concepts, and analyze its advantages in more real-world downstream tasks, like semantic navigation.
We plan to add stronger post-processing to ensure graph expansion stays focused given the lexical variations observed during evaluation (see~\cref{fig:example_subg_hierachy}).
Additionally, we aim to open source the pipeline code and related datasets until the final paper publication.

\printbibliography[title=References]
\end{document}